\documentclass[runningheads]{llncs}
\usepackage{multirow}
\usepackage{booktabs}
\usepackage[table]{xcolor}
\usepackage{orcidlink}
\usepackage{amsmath}
\usepackage{array}
\usepackage{hyperref}
\usepackage[T1]{fontenc}
\usepackage{makecell}
\usepackage{graphicx,verbatim}
\usepackage{color}

\begin{document}
\title{Mitigating Data Exfiltration Attacks through Layer-Wise Learning Rate Decay Fine-Tuning}
\titlerunning{Mitigating Data Exfiltration via Layer-Wise LR Decay FT}

\author{
Elie Thellier\orcidlink{0009-0007-2747-1637}\textsuperscript{*} \and
Huiyu Li \and
Nicholas Ayache \and
Hervé Delingette\orcidlink{0000-0001-6050-5949}
}


\authorrunning{E. Thellier et al.}

\institute{
Centre Inria d'Université Côte d'Azur, Epione Team, Sophia Antipolis, France\\
\textsuperscript{*}Corresponding author: \email{elie.thellier@inria.fr}
}

\maketitle

\begin{abstract}
Data lakes enable the training of powerful machine learning models on sensitive, high-value medical datasets, but also introduce serious privacy risks due to potential leakage of protected health information. Recent studies show adversaries can exfiltrate training data by embedding latent representations into model parameters or inducing memorization via multi-task learning. These attacks disguise themselves as benign utility models while enabling reconstruction of high-fidelity medical images, posing severe privacy threats with legal and ethical implications. In this work, we propose a simple yet effective mitigation strategy that perturbs model parameters at export time through fine-tuning with a decaying layer-wise learning rate to corrupt embedded data without degrading task performance. Evaluations on DermaMNIST, ChestMNIST, and MIMIC-CXR show that our approach maintains utility task performance, effectively disrupts state-of-the-art exfiltration attacks, outperforms prior defenses, and renders exfiltrated data unusable for training. Ablations and discussions on adaptive attacks highlight challenges and future directions. Our findings offer a practical defense against data leakage in data lake-trained models and centralized federated learning.


\end{abstract}

\section{Introduction}

To develop better AI models for medical data processing, hospitals and other data owners are creating medical data lakes \cite{gentner2023data}. These infrastructures provide controlled remote access to rare, privacy-critical data, such as dermatoscopic or x-ray images. Access to these systems is strictly regulated, as any leakage of sensitive medical information could pose a serious reputational risk for data owners and create re-identification threats for individuals \cite{kaissis2020secure}. Despite these safeguards, recent studies have revealed significant vulnerabilities in current defense mechanisms \cite{gong2024hidden}, highlighting the urgent need for effective mitigation strategies. 

Defense mechanisms must consider that models trained on sensitive datasets can memorize dataset properties, intentionally or unintentionally, through attacks like property inference \cite{wu2020evaluation}, membership inference \cite{xu2023membership}, model inversion \cite{dibbo2023sok}, backdoor attacks \cite{luzon2024memory}, or simply overfitting \cite{yeom2018privacy}. In particular, \textbf{data exfiltration attacks} enable models to memorize and leak raw training data. Recent state-of-the-art methods such as Transpose \cite{amit2023transpose} and DEC \cite{li2022data} highlight this risk by using neural networks as covert containers to exfiltrate data from protected environments. Transpose employs a reversible deep network to secretly memorize images while appearing to perform legitimate classification. DEC compresses target data via a pre-trained encoder, embedding it steganographically into a utility model for later extraction and reconstruction. Additionally, diffusion models have been shown to memorize and leak training data through content extraction and membership inference \cite{carlini2023extracting}.

In response to these threats, several best practices have been proposed to protect data lakes against data theft, including differential privacy \cite{adnan2022federated}, model watermarking \cite{giakoumaki2006secure}, and manual model inspection. However, these approaches often struggle to balance performance, robustness, and computational cost. More targeted defenses like Fine-Pruning \cite{liu2018fine} and Super-Fine-Tuning \cite{sha2022fine} address specific attacks such as backdoors, but the defense landscape remains fragmented. Broader strategies such as training on synthetic data or strong anonymization have also been explored. For example, \cite{li2025generativemedicalimageanonymization} implements medical image anonymization by disentangling utility and identity in latent representations and \cite{seo2024generative} introduces identity unlearning to prevent generative models from reproducing individuals. Despite these advances, a key gap remains: to our knowledge, no prior work has specifically evaluated mitigation strategies designed for neural network-based data exfiltration attacks. 

In this paper, we fill this gap by introducing a straightforward but effective method to reduce neural networks' memorization ability while keeping their performance largely intact. Our approach is based on a fine-tuning protocol that applies a decaying, layer-wise learning rate, to disrupt the early layers of the model while preserving the stability of the output layers. We evaluate our method against two state-of-the-art data exfiltration attacks, Transpose and an improved version of DEC, using several medical datasets of various resolutions. In addition, we benchmark our approach against existing mitigation baselines, design a usability test, present a detailed ablation study and explore the impact of adaptive adversaries. Those experiments show the good trade-off between privacy and accuracy that one can reach using our novel fine-tuning mitigation method. While our method is designed for post-training sanitization of models exported from centralized data lakes, a similar risk arises in centralized federated learning when local models trained on sensitive data are shared with a central server \cite{li2022data}. In such cases, our approach can be applied at the point of model export or aggregation to mitigate data exfiltration.

\subsection{Related Mitigation Methods}

This section presents fine-tuning-based defenses designed to mitigate data stealing while preserving classification utility. All baselines are then implemented with hyperparameters sourced from original papers or tuned by us.

\textbf{Vanilla Fine-Tuning (Vanilla FT)} retrains the full model on the utility task using original hyperparameters for 3 to 10 epochs (depending on dataset size). The goal is to adapt model weights for utility while reducing memorization: \(\min_{\theta} \mathcal{L}_{ut}(\theta; X, y)\) starting from trained \( \theta_0 \), with gradient descent updates: \(\theta_{t+1} = \theta_t - \eta \nabla_\theta \mathcal{L}_{ut}(\theta_t)\), with \(\eta = \eta_{training}= 1 \times 10^{-4} \). A variation of this approach is \textbf{High LR Fine-Tuning (High LR FT)}, which uses a 100 $\times$ higher learning rate \( \left( \eta = 1 \times 10^{-2}\right) \) to escape memorization-related local minima, while optimizing for utility. Extending these elemental fine-tuning methods,  \textbf{Super-Fine-Tuning \cite{sha2022fine} (Super-FT)} employs cyclical learning rates to alternate between disruption and recovery phases. The learning rate at step $t$ follows: \(\eta(t) = \eta_\text{base} + \left(1 - \left|2 \frac{t \bmod C}{C} - 1\right|\right) \cdot (\eta_\text{MAX} - \eta_\text{base})\), where \(\eta_\text{base} = 1 \times 10^{-4}\), $C$ is the cycle length, and $\eta_\text{MAX}$ is \(1 \times 10^{-1}\) (in phase 1) or \(1 \times 10^{-3}\) (in phase 2, starting after 10\% of training). In contrast, \textbf{Weight Decay Fine-Tuning (WD FT)} adds L2 regularization to discourage large weights, which are often associated with memorization: \(\theta_{t+1} = \theta_t - \eta \left( \nabla_\theta \mathcal{L}_{ut}(\theta_t) + \lambda \theta_t \right)\) where $\lambda = 10^{-2}$ is the weight decay coefficient. 

Other techniques consist of streamlined weight modifications before a fine-tuning step. \textbf{Random Weight Perturbation (RWP)} injects Gaussian noise to parameters before fine-tuning to disrupt memorized patterns then restore task accuracy: \(\theta' = \theta + \epsilon, \quad \epsilon \sim \mathcal{N}(0, \sigma^2)\) where $\sigma = 10^{-2}$ controls noise magnitude. \textbf{Fine-Pruning \cite{liu2018fine}} adopts a structural approach by removing small-magnitude weights within the last convolutional layer (allowing up to 4\% accuracy drop), then fine-tunes the masked model. Specifically, $\theta' = \theta \odot m$, where \( m_i = 0 \) if $|\theta_i| \leq \tau$ with \(m\) as a binary mask and \(\tau\) the pruning threshold. Utility is recovered by optimizing over remaining weights: \(\min_{\theta'} \mathcal{L}_{ut}(\theta'; X, y)\). Similarly, \textbf{Random Weight Dropout (RWD)} applies random independent binary masks to zero out weights with probability \( p \): \(\theta_i' = \theta_i \cdot z_i, \quad z_i \sim \text{Bernoulli}(1 - p)\) and fine-tunes the model to restore task performance.

Finally, \textbf{Transpose Detection \cite{amit2023transpose}} is a detection-only method targeting Transpose-type attacks. It optimizes a latent code to test whether the transposed model can reconstruct training-like data, success indicates memorization. While effective against Transpose attacks, it requires manual model transposition and does not provide mitigation.

\section{Layer-Wise Learning Rate Decay Fine-Tuning}

\begin{figure}[htbp]
  \centering
  \includegraphics[width=0.9\linewidth]{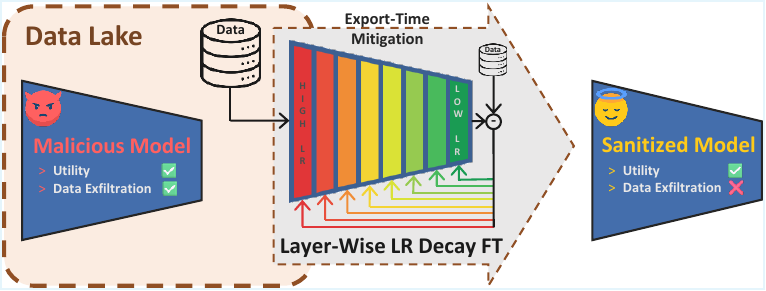}
  \caption{Overview of our export-time mitigation. A malicious model trained on the data lake may retain both utility and private data. At export, we apply LWLRD FT using training data to disrupt early-layer memorization while preserving task performance. The sanitized model retains utility but loses reconstruction ability.}
  \label{fig:method}
\end{figure}

We introduce a novel fine-tuning mitigation approach building on strategies like LARS \cite{ginsburg2018large} and AutoLR \cite{ro2021autolr}, which assign varying learning rates across layers to enhance training. Traditional Layer-wise Learning Rate Decay, widely used in NLP and ViT, fine-tunes task-specific layers with higher rates while preserving low-level features \cite{dong2022clip}. 

In contrast, our Layer-Wise Learning Rate Decay Fine-Tuning (LWLRD FT) method deliberately reverses this strategy by assigning higher learning rates to early layers, where memorization tends to occur, to disrupt memorization, and lower learning rates to later layers, where task-relevant features reside, to preserve their utility (see Fig.~\ref{fig:method}).
The per layer learning rate \(\eta_\ell\) mapping is defined as: \( \eta_\ell = \eta_{\text{high}} \cdot \left(\frac{\eta_{\text{low}}}{\eta_{\text{high}}}\right)^{\frac{\ell - 1}{L - 1}} \) where \( L \) is the total number of layers, \( \ell \in [1, L] \) is the layer index (with \(\ell = 1\) at the first), \( \eta_{\text{high}} = 1 \times 10^{-2}\) is the learning rate for the first layer, and \( \eta_{\text{low}} = 1 \times 10^{-4}\) for the last. 

LWLRD FT is especially effective against threats like Transpose and DEC, which exploit early-layer stability to reconstruct or hide training data. DEC encodes sensitive content via steganography in early-layer parameters, while Transpose models repurpose early layers as a reconstruction head because of the model inversion. By targeting updates to these early layers, LWLRD FT offers a lightweight yet effective export-time mitigation, approaching the benefits of full retraining with significantly reduced overhead and low-to-no loss in utility.

\section{Experimental Setup}

In our scenario (shown in Fig.~\ref{fig:method}), a data exfiltration attack produces a malicious model within the data lake capable of reconstructing sensitive medical images. To mitigate this, we apply our mitigation method during export, using only the original training data, hyperparameters and additional method-dependent settings. The resulting sanitized model is no longer able to reconstruct images.

\subsection{Attacks}

We focus on two prominent attacks: \textbf{Transpose} \cite{amit2023transpose} and \textbf{Data Exfiltration by Compression (DEC)} \cite{li2022data}. For comparison, we limit the number of extracted samples to 1,000 (and 100 for Transpose on MIMIC-CXR), although, both attacks can recover more images depending on model and data size.

\textbf{The Transpose attack} trains a single model simultaneously on two tasks: a visible utility task and a hidden memorization task, by running the model in both directions. Yet, this malicious approach introduces implementation challenges. First, some layers are mathematically irreversible, thus not transposable, limiting the quality of memorized images. Second, the current implementation restricts the attack's utility task to classification problems only. Finally, the two-tasks learning procedure tends to restrain classification convergence in favor of memorization, thereby limiting the classification performance of Transpose models. We implement the model transposition and training following \cite{amit2023transpose}, using two AdamW optimizers \cite{loshchilov2019decoupled} with learning rates of $1 \times 10^{-4}$ for classification and $1 \times 10^{-3}$ for memorization, alongside a learning rate scheduler for memorization.

\textbf{The DEC attack} employs a multi-task learning framework composed of a compression network and a separate utility branch, making DEC adaptable to segmentation or detection. However, vanilla DEC is limited by the output size of its HiFiC compression network and by a fragile steganography technique. To improve its reliability, we replace the compressor with an AE-GAN \cite{li2025generativemedicalimageanonymization} trained on external data. The encoder converts target data into compact 512-length latent vectors in the data lake, while the utility network is trained on the classification task, using AdamW \cite{loshchilov2019decoupled} at a learning rate of $1 \times 10^{-4}$, concealing the attack. Latent codes are hidden via steganography in the 16 least significant bits of the utility network's float32 parameters using a custom 16-bit format that shifts by 1, scales by 20,000, and rounds floats to integers before bit-encoding. This restricts values to the latent code range [-0.5, 2.5], matching the AE-GAN distribution, rather than the IEEE 754 float32 range, improving robustness by preventing value explosions. Finally, the model is exported, latent codes are extracted, and the AE-GAN generator decodes them to reconstruct the stolen data.

\subsection{Datasets, Models, and Metrics}

We evaluate our mitigation and baseline methods on three medical imaging datasets. \textbf{DermaMNIST}~\cite{medmnistv1,dermamnist1,dermamnist2} is a 7-class classification task with 10,015 low-resolution ($28\times28$) images, using a \textbf{ResNet18} \cite{he2016deep} model, leading to a latent capacity for DEC of 21,820. While its clinical relevance is limited, DermaMNIST offers a reproducible benchmark for trend analysis and hyperparameter tuning. \textbf{ChestMNIST} \cite{medmnistv2,chestmnist} involves 14-label classification with 112,120 images of size $224\times224$, also using \textbf{ResNet18}. \textbf{MIMIC-CXR} \cite{johnson2019mimic} is a higher-resolution ($512\times512$) 4-label classification task with 54,038 samples, using \textbf{DenseNet121} \cite{huang2017densely}, with a DEC latent capacity of 14,612. These models are selected for their strong reported utility baselines \cite{medmnistv1,medmnistv2,li2024data}, enabling reliable comparisons.

We evaluate mitigation methods by their effect on model utility, using average AUC and label-wise accuracy (acc) computed on the original test set. To quantify data leakage and assess the model’s image reconstruction capability, we use SSIM \cite{wang2004image}, LPIPS \cite{zhang2018unreasonable} and PSNR between original training images and their stolen reconstructions. The goal is to preserve classification performance while degrading reconstruction quality. We also report mitigation duration to verify practical deployment at export. All experiments run on an H100 NVL GPU, with results averaged over multiple runs for statistical robustness. 

To further evaluate privacy, we conduct a usability test simulating a practical attack scenario: an adversary reconstructs training data and is also assumed to have access to corresponding labels, then trains a new classifier from scratch on this stolen dataset. We evaluate the classifier on the original task’s test set to assess whether the exfiltrated data retains enough task-relevant information to support model training, something mitigation should ideally prevent.

\begin{table}[htbp]
\centering
\caption{Comparison of mitigation methods under the Transpose attack. Best values are bolded, second-best underlined. auc, acc and ssim are reported in \%.}
\label{tab:transpose_mitigation}
{\fontsize{8}{9.6}\selectfont
\setlength{\tabcolsep}{1.9pt}
\begin{tabular}{@{} lcccc|ccccc|ccccc @{}}
\toprule
\multirow{2}{*}{\textbf{Method}} 
& \multicolumn{4}{c|}{\textbf{DermaMNIST}} 
& \multicolumn{5}{c|}{\textbf{ChestMNIST}} 
& \multicolumn{5}{c}{\textbf{MIMIC-CXR}} \\
\cmidrule(lr){2-5} \cmidrule(lr){6-10} \cmidrule(l){11-15}
& auc & acc & ssim & psnr
& auc & acc & ssim & psnr & lpips
& auc & acc & ssim & psnr & lpips \\
\midrule
\rowcolor{gray!15}
No Mitigation      & 88.1 & 72.7 & 95.6 & 37.6
                   & 67.9 & 91.9 & 85.4 & 30.4 & 0.24
                   & 49.7 & 64.6 & 71.2 & 23.8 & 0.46 \\
\midrule
Vanilla FT         & 86.9 & \textbf{71.4} & 73.2 & 17.7     
                   & 67.9 & 93.3 & 61.1 & 20.5 & 0.37
                   & 59.0 & 56.9 & 54.2 & 16.6 & 0.52 \\
\rowcolor{gray!10}
High LR FT         & 85.9 & 68.9 & 13.6 & \underline{6.3}    
                   & 67.0 & 93.2 & 23.0 & \textbf{5.0} & \underline{0.72}   
                   & 50.0 & 53.9 & \underline{30.5} & \textbf{4.9} & \underline{0.66} \\
Super-FT           & 85.0 & 68.8 & \underline{12.1} & \textbf{5.9}    
                   & 66.5 & \textbf{94.7} & 31.5 & \underline{5.3} & 0.64 
                   & 50.0 & \textbf{60.9} & 42.0 & \underline{5.0} & 0.62 \\
\rowcolor{gray!10}
WD FT              & 86.8 & 70.9 & 67.3 & 15.9     
                   & 67.9 & \underline{93.4} & 59.7 & 20.1 & 0.36
                   & 59.0 & 56.0 & 54.2 & 16.6 & 0.52 \\
RWP + FT           & \underline{87.1} & \underline{71.1} & 39.0 & 15.1     
                   & 67.8 & 93.3 & \textbf{20.4} & 14.8 & 0.49  
                   & 59.6 & 56.5 & 53.2 & 16.4 & 0.53 \\
\rowcolor{gray!10}
Fine-Pruning       & 87.0 & 70.9 & 38.1 & 12.4    
                   & \underline{71.4} & \textbf{94.7} & 55.1 & 14.5 & 0.44   
                   & \underline{60.7} & 53.3 & 49.5 & 14.6 & 0.53 \\
RWD + FT           & 86.9 & \underline{71.1} & 58.0 & 13.4     
                   & 68.1 & \underline{93.4} & 43.8 & 17.0 & 0.43
                   & 59.3 & \underline{57.0} & 50.2 & 15.7 & 0.54 \\
\rowcolor{gray!15}
\textbf{LWLRD FT}  & \textbf{87.8} & 70.5 & \textbf{11.5} & \textbf{5.9}    
                   & \textbf{74.3} & \textbf{94.7} & \underline{20.7} & \textbf{5.0} & \textbf{0.74}
                   & \textbf{66.2} & 55.3 & \textbf{13.3} & 5.1 & \textbf{0.80} \\
\bottomrule
\end{tabular}
}
\end{table}

\section{Results}

\subsection{Performance Against the Transpose Attack}

\begin{figure}[!htbp]
  \centering
  \includegraphics[width=\linewidth]{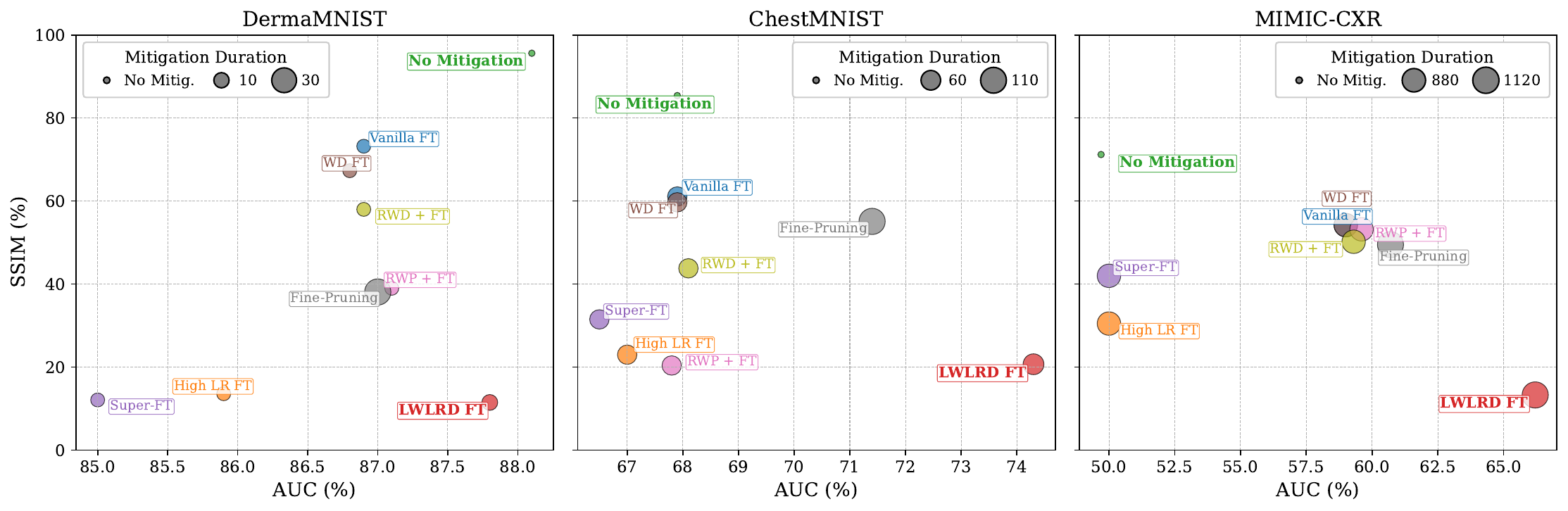}
  \caption{Mitigation methods against the Transpose attack. Points show AUC (utility) vs. SSIM (leakage); lower-right is better. Size represents mitigation time.}
  \label{fig:pca1}
\end{figure}

\begin{figure}[!htbp]
  \centering
  \includegraphics[width=\linewidth]{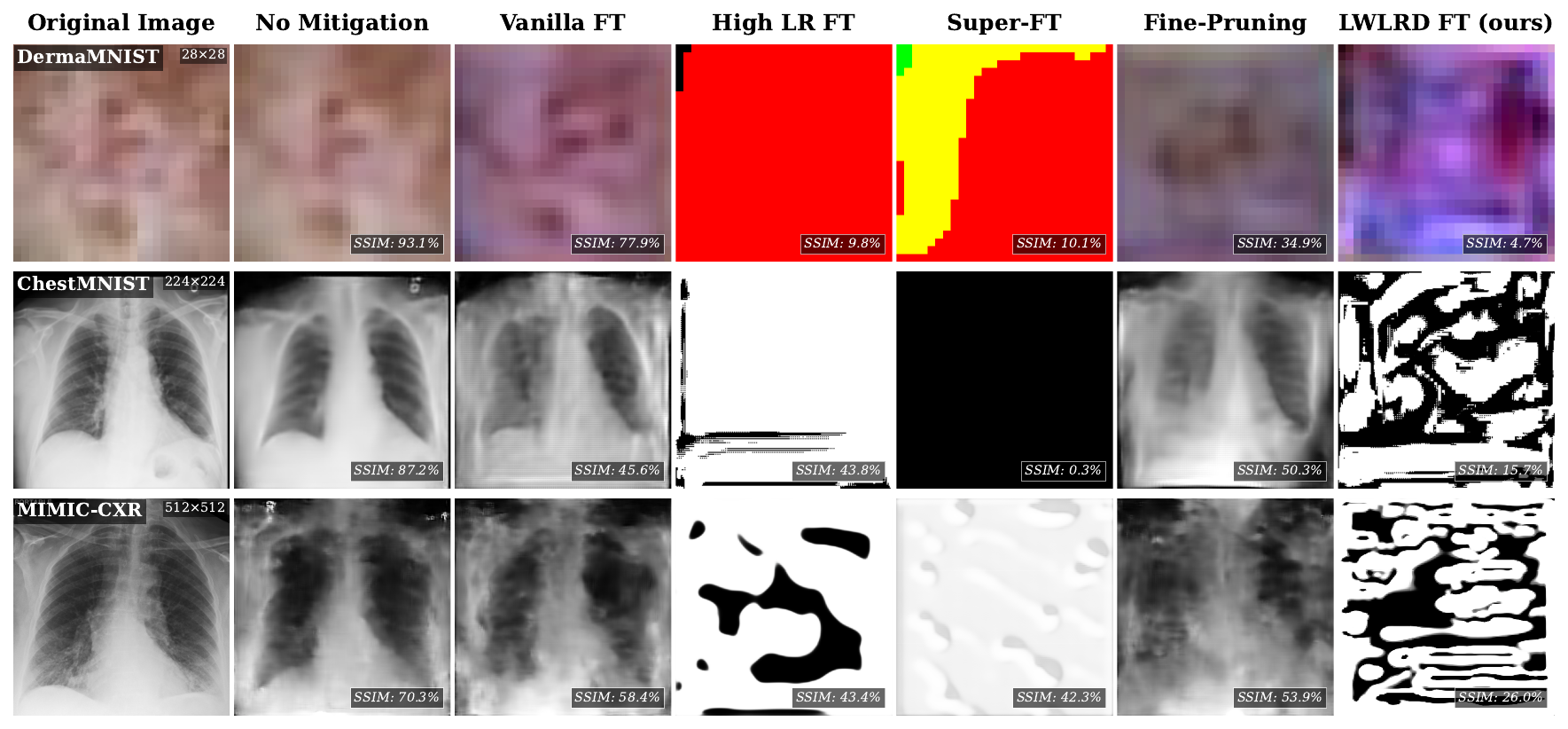}
  \caption{Transpose reconstructions by mitigation method. Each row shows the original image (left) and mitigated reconstructions with SSIM scores.}
  \label{fig:qualitative}
\end{figure}

Table~\ref{tab:transpose_mitigation} summarizes the performance of our mitigation and baselines against the Transpose attack. \textbf{LWLRD FT} achieves the strongest privacy protection (lowest SSIM and PSNR, highest LPIPS) while maintaining competitive or superior utility compared to baselines. The lower initial accuracy on ChestMNIST and MIMIC-CXR results from conflicting multi-task learning and limited convergence. The results demonstrate the effectiveness of LWLRD FT at corrupting exfiltrated images without sacrificing task performance, with mitigation time 27\% longer than Vanilla FT due to the layer-wise learning rate mapping. Super-Fine-Tuning and Fine-Pruning offer reasonable defense but often reduce utility or leave higher-quality reconstructions intact; Vanilla FT is insufficient. Fig.~\ref{fig:pca1} visualizes the privacy-utility trade-off and includes mitigation duration, while Fig.~\ref{fig:qualitative} presents qualitative reconstruction examples before and after mitigation.

\subsection{Performance Against the DEC Attack}

\begin{figure}[!htbp]
  \centering
  \includegraphics[width=\linewidth]{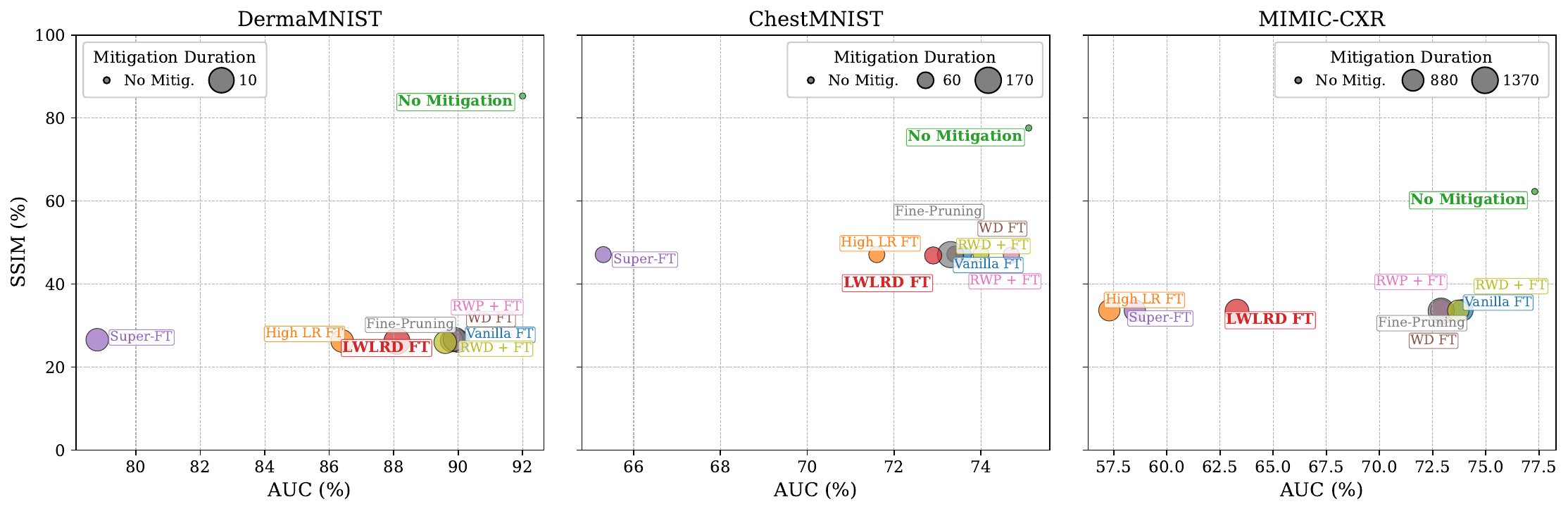}
  \caption{Mitigation methods against the DEC attack. Points represent utility versus leakage; leakage decreases similarly across methods due to DEC’s fragile steganography, while utility varies. Point size indicates mitigation duration.}
  \label{fig:pca2}
\end{figure}

Fig.~\ref{fig:pca2} shows similar mitigation results against the DEC attack across methods, as the steganographic latent code is fragile: small bit flips or noise easily destroy the hidden data. Our adapted DEC, using a custom bit representation, reconstructs latent codes within the AE-GAN range from perturbed parameters, producing visibly distorted images and ensuring strong privacy regardless of mitigation method. Consequently, privacy metrics show minimal variation, with all methods effectively neutralizing DEC, though accuracy differs. Notably, LWLRD FT fails to recover utility on MIMIC-CXR, likely due to DenseNet121's complex dense layers being more sensitive to disruption than ResNet18's simpler structure.

\subsection{Usability Test}

Even when reconstructed images become visually unrecognizable after mitigation, attackers with stolen labels can sometimes train models that perform better than random guessing. Without mitigation, models trained on stolen data reach AUCs of 81.7\% (DermaMNIST), 56.0\% (ChestMNIST), and 50.8\% (MIMIC-CXR), reflecting some retained task information, though low performance likely results from the limited amount of stolen data. With LWLRD FT, these models achieve 51.1\% AUC (DermaMNIST) and below 50\% (ChestMNIST, MIMIC-CXR), indicating the attacker’s classifier is essentially guessing. This shows our mitigation effectively degrades the usability of stolen data for model training.




\subsection{Ablation Study}

\begin{figure}[!htbp]
  \centering
  \includegraphics[width=\linewidth]{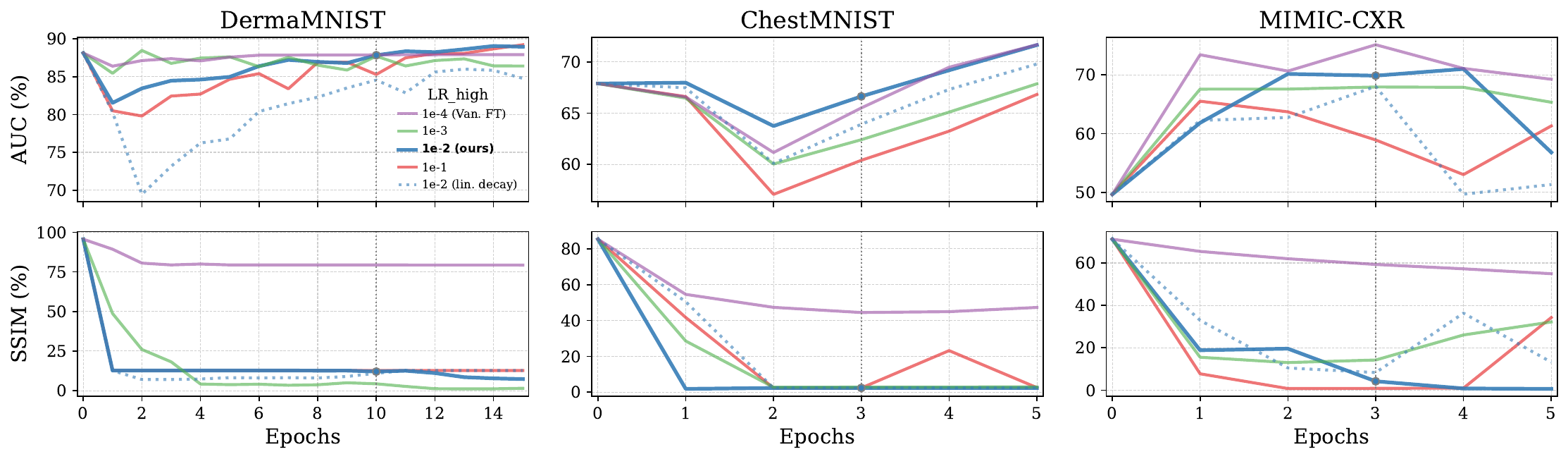}
  \caption{LWLRD FT hyperparameter study under Transpose Attack. AUC and SSIM over epochs for varying \(\eta_{\text{high}}\) and decay (solid: exponential, dashed: linear).}
  \label{fig:abla}
\end{figure}

We conducted an ablation study on LWLRD FT hyperparameters using the Transpose attack, examining fine-tuning duration, early-layer learning rate \(\eta_{\text{high}}\), and decay strategy. Fig.~\ref{fig:abla} shows that exponential decay restores classification performance faster than linear decay. Longer fine-tuning improves utility recovery but increases cost, while reconstruction is disrupted early. The learning rate \(\eta_{\text{high}} = 1 \times 10^{-2}\) provides the best balance of effectiveness and stability.

\section{Discussion and Conclusion}

Our novel mitigation method disrupts current data exfiltration attacks, though adaptive techniques may improve resilience. The DEC attack produces high-fidelity reconstructions but depends on fragile steganography. Our custom float-to-bit encoding is not reversible, causing precision loss even without mitigation, which degrades latent codes into noise. More robust or error-correcting encodings could enhance recovery. As a stronger alternative to LSB steganography, we propose learned steganography, in which attackers initialize and freeze early weights with latent codes during utility training, then unfreeze later to avoid suspicion. Preliminary tests suggest it offers better image recovery and mitigation robustness, though it may be more vulnerable to random weight dropout. 

To conclude, we introduced a fine-tuning strategy that perturbs early model layers during export to disrupt embedded data leakage without impacting performance. This method reduces data exfiltration success across the evaluated medical imaging datasets and shows advantages over existing mitigations. We recommend data lake owners consider this adaptive fine-tuning before model export as a practical defense against covert data theft. Similar mitigation may be applied in centralized federated learning when sharing or aggregating local models. During mitigation, performance should be monitored, with hyperparameters adjusted to preserve accuracy. Future work should extend these defenses to distributed training and address adaptive threats like learned steganography.

\begin{credits}
\subsubsection{\ackname} This work was supported by the Région PACA and France 2030 initiative through the funding of the I-Démo project PLICIA, and by the French government through the National Research Agency (ANR), under the IA Cluster projects (ANR-23-IACL-0001). Experiments in this paper were carried out using the Abaca Cluster, part of the Grid'5000 testbed, supported by a scientific interest group hosted by Inria and including CNRS, RENATER, several universities, and other organizations.

\subsubsection{\discintname} The authors have no competing interests to declare that are relevant to the content of this article. 

\end{credits}

%
%
%
\bibliographystyle{splncs04-num} 
\bibliography{Paper-0002}

\begin{thebibliography}{10}
\providecommand{\url}[1]{\texttt{#1}}
\providecommand{\urlprefix}{URL }
\providecommand{\doi}[1]{https://doi.org/#1}

\bibitem{gentner2023data}
Gentner, T., Neitzel, T., Schulze, J., Gerschner, F., Theissler, A.: Data lakes
  in healthcare: applications and benefits from the perspective of data sources
  and players. Procedia Computer Science  \textbf{225},  1302--1311 (2023)

\bibitem{kaissis2020secure}
Kaissis, G.A., Makowski, M.R., R{\"u}ckert, D., Braren, R.F.: Secure,
  privacy-preserving and federated machine learning in medical imaging. Nature
  Machine Intelligence  \textbf{2}(6),  305--311 (2020)

\bibitem{gong2024hidden}
Gong, X., Wang, Y., Li, S., Sun, M., Li, S., Wang, Q., Lam, K.Y., Chen, C.:
  Hidden data privacy breaches in federated learning. arXiv preprint
  arXiv:2411.18269  (2024)

\bibitem{wu2020evaluation}
Wu, M., Zhang, X., Ding, J., Nguyen, H., Yu, R., Pan, M., Wong, S.T.:
  Evaluation of inference attack models for deep learning on medical data.
  arXiv preprint arXiv:2011.00177  (2020)

\bibitem{xu2023membership}
Xu, T., Liu, C., Zhang, K., Zhang, J.: Membership inference attacks against
  medical databases. In: International Conference on Neural Information
  Processing. pp. 15--25. Springer (2023)

\bibitem{dibbo2023sok}
Dibbo, S.V.: Sok: Model inversion attack landscape: Taxonomy, challenges, and
  future roadmap. In: 2023 IEEE 36th Computer Security Foundations Symposium
  (CSF). pp. 439--456. IEEE (2023)

\bibitem{luzon2024memory}
Luzon, E., Amit, G., Weiss, R., Mirsky, Y.: Memory backdoor attacks on neural
  networks. arXiv preprint arXiv:2411.14516  (2024)

\bibitem{yeom2018privacy}
Yeom, S., Giacomelli, I., Fredrikson, M., Jha, S.: Privacy risk in machine
  learning: Analyzing the connection to overfitting. In: 2018 IEEE 31st
  computer security foundations symposium (CSF). pp. 268--282. IEEE (2018)

\bibitem{amit2023transpose}
Amit, G., Levy, M., Mirsky, Y.: Transpose attack: Stealing datasets with
  bidirectional training. arXiv preprint arXiv:2311.07389  (2023)

\bibitem{li2022data}
Li, H., Ayache, N., Delingette, H.: Data stealing attack on medical images: Is
  it safe to export networks from data lakes? In: International Workshop on
  Distributed, Collaborative, and Federated Learning. pp. 28--36. Springer
  (2022)

\bibitem{carlini2023extracting}
Carlini, N., Hayes, J., Nasr, M., Jagielski, M., Sehwag, V., Tramer, F., Balle,
  B., Ippolito, D., Wallace, E.: Extracting training data from diffusion
  models. In: 32nd USENIX Security Symposium (USENIX Security 23). pp.
  5253--5270 (2023)

\bibitem{adnan2022federated}
Adnan, M., Kalra, S., Cresswell, J.C., Taylor, G.W., Tizhoosh, H.R.: Federated
  learning and differential privacy for medical image analysis. Scientific
  reports  \textbf{12}(1), ~1953 (2022)

\bibitem{giakoumaki2006secure}
Giakoumaki, A., Pavlopoulos, S., Koutsouris, D.: Secure and efficient health
  data management through multiple watermarking on medical images. Medical and
  Biological Engineering and Computing  \textbf{44},  619--631 (2006)

\bibitem{liu2018fine}
Liu, K., Dolan-Gavitt, B., Garg, S.: Fine-pruning: Defending against
  backdooring attacks on deep neural networks. In: International symposium on
  research in attacks, intrusions, and defenses. pp. 273--294. Springer (2018)

\bibitem{sha2022fine}
Sha, Z., He, X., Berrang, P., Humbert, M., Zhang, Y.: Fine-tuning is all you
  need to mitigate backdoor attacks. arXiv preprint arXiv:2212.09067  (2022)

\bibitem{li2025generativemedicalimageanonymization}
Li, H., Ayache, N., Delingette, H.: Generative medical image anonymization
  based on latent code projection and optimization. In: 2025 IEEE 22nd
  International Symposium on Biomedical Imaging (ISBI). pp.~1--4. IEEE (2025)

\bibitem{seo2024generative}
Seo, J., Lee, S.H., Lee, T.Y., Moon, S., Park, G.M.: Generative unlearning for
  any identity. In: Proceedings of the IEEE/CVF Conference on Computer Vision
  and Pattern Recognition. pp. 9151--9161 (2024)

\bibitem{ginsburg2018large}
Ginsburg, B., Gitman, I., You, Y.: Large batch training of convolutional
  networks with layer‑wise adaptive rate scaling. In: ICLR\,2018 Conference
  (2018)

\bibitem{ro2021autolr}
Ro, Y., Choi, J.Y.: Autolr: Layer‑wise pruning and auto‑tuning of learning
  rates in fine‑tuning of deep networks. In: Proceedings of the AAAI
  Conference on Artificial Intelligence (AAAI‑21). pp. 2486--2494 (2021)

\bibitem{dong2022clip}
Dong, X., Bao, J., Zhang, T., Chen, D., Gu, S., Zhang, W., Yuan, L., Chen, D.,
  Wen, F., Yu, N.: Clip itself is a strong fine-tuner: Achieving 85.7\% and
  88.0\% top-1 accuracy with vit-b and vit-l on imagenet. arXiv preprint
  arXiv:2212.06138  (2022)

\bibitem{loshchilov2019decoupled}
Loshchilov, I., Hutter, F.: Decoupled weight decay regularization. In:
  International Conference on Learning Representations (ICLR) (2019)

\bibitem{medmnistv1}
Yang, J., Shi, R., Ni, B.: Medmnist classification decathlon: A lightweight
  automl benchmark for medical image analysis. In: IEEE 18th International
  Symposium on Biomedical Imaging (ISBI). pp. 191--195 (2021)

\bibitem{dermamnist1}
Tschandl, P., Rosendahl, C., Kittler, H.: The ham10000 dataset, a large
  collection of multi-source dermatoscopic images of common pigmented skin
  lesions. Scientific data p. 180161 (2018)

\bibitem{dermamnist2}
Codella, N., Rotemberg, V., Tschandl, P., Celebi, M.E., Dusza, S., Gutman, D.,
  Helba, B., Kalloo, A., Liopyris, K., Marchetti, M., et~al.: Skin lesion
  analysis toward melanoma detection 2018: A challenge hosted by the
  international skin imaging collaboration (isic). arXiv preprint
  arXiv:1902.03368  (2019)

\bibitem{he2016deep}
He, K., Zhang, X., Ren, S., Sun, J.: Deep residual learning for image
  recognition. In: Proceedings of the IEEE Conference on Computer Vision and
  Pattern Recognition (CVPR). pp. 770--778 (2016)

\bibitem{medmnistv2}
Yang, J., Shi, R., Wei, D., Liu, Z., Zhao, L., Ke, B., Pfister, H., Ni, B.:
  Medmnist v2-a large-scale lightweight benchmark for 2d and 3d biomedical
  image classification. Scientific Data  \textbf{10}(1), ~41 (2023)

\bibitem{chestmnist}
Wang, X., Peng, Y., et~al.: Chestx-ray8: Hospital-scale chest x-ray database
  and benchmarks on weakly-supervised classification and localization of common
  thorax diseases. In: CVPR. pp. 3462--3471 (2017)

\bibitem{johnson2019mimic}
Johnson, A.E., Pollard, T.J., Greenbaum, N.R., Lungren, M.P., Deng, C.y., Peng,
  Y., Lu, Z., Mark, R.G., Berkowitz, S.J., Horng, S.: Mimic-cxr-jpg, a large
  publicly available database of labeled chest radiographs. arXiv preprint
  arXiv:1901.07042  (2019)

\bibitem{huang2017densely}
Huang, G., Liu, Z., Van Der~Maaten, L., Weinberger, K.Q.: Densely connected
  convolutional networks. In: Proceedings of the IEEE Conference on Computer
  Vision and Pattern Recognition (CVPR). pp. 4700--4708 (2017)

\bibitem{li2024data}
Li, H.: Data exfiltration and anonymization of medical images based on
  generative models. Ph.D. thesis, Universit{\'e} C{\^o}te d'Azur (2024)

\bibitem{wang2004image}
Wang, Z., Bovik, A.C., Sheikh, H.R., Simoncelli, E.P.: Image quality
  assessment: From error visibility to structural similarity. IEEE Transactions
  on Image Processing  \textbf{13}(4),  600--612 (2004)

\bibitem{zhang2018unreasonable}
Zhang, R., Isola, P., Efros, A.A., Shechtman, E., Wang, O.: The unreasonable
  effectiveness of deep features as a perceptual metric. In: Proceedings of the
  IEEE conference on computer vision and pattern recognition. pp. 586--595
  (2018)

\end{thebibliography}
\end{document}